\documentclass{article}
\pdfoutput=1
\usepackage{spconf,graphicx} 

\usepackage{cite}
\usepackage{amsmath,amssymb,amsfonts}
\usepackage{algorithm}
\usepackage{algpseudocode}
\usepackage{mathtools}

\usepackage{enumitem}
\setlist{nosep, leftmargin=14pt}

\usepackage{mwe} % to get dummy images
\usepackage{booktabs} 
\usepackage{multirow}
\usepackage{tabularx}
\usepackage{adjustbox}
\usepackage{array}
\usepackage{svg}
\usepackage{gensymb}
\usepackage{anyfontsize}
\usepackage{xcolor}
\usepackage{soul}
\usepackage{scalerel}
\usepackage{stackengine}

\usepackage{amssymb}% http://ctan.org/pkg/amssymb
\usepackage{pifont}% http://ctan.org/pkg/pifont
\newcommand{\cmark}{\ding{51}}%
\usepackage[colorlinks=true,linkcolor=red,citecolor=blue, urlcolor=blue]{hyperref}

\title{A Point in the Right Direction: Vector Prediction for Spatially-aware Self-supervised Volumetric Representation Learning}
\name{Yejia Zhang$^{\dagger}$ 
     \quad Pengfei Gu$^{\dagger}$ 
     \quad Nishchal Sapkota$^{\dagger}$
     \quad Hao Zheng$^{\dagger}$ 
     \quad Peixian Liang$^{\dagger}$ 
     \quad Danny Z. Chen$^{\dagger}$ 
     }
\address{ $^{\dagger}$University of Notre Dame, Department of Computer Science and Engineering, Notre Dame, IN, USA}

\begin{document}
%\ninept
%

\maketitle

\begin{abstract}
% 100 - 150 words, identical to text-submitted one || 
High annotation costs and limited labels for dense 3D medical imaging tasks have recently motivated an assortment of 3D self-supervised pretraining methods that improve transfer learning performance. 
However, these methods commonly lack spatial awareness despite its centrality in enabling effective 3D image analysis.
More specifically, position, scale, and orientation are not only informative but also automatically available when generating image crops for training.
Yet, to date, no work has proposed a pretext task that distills all key spatial features.
%Given these insights, 
To fulfill this need, we develop a new self-supervised method, \textbf{VectorPOSE}, which promotes better spatial understanding with two novel pretext tasks: Vector Prediction (\textbf{VP}) and Boundary-Focused Reconstruction (\textbf{BFR}).
VP focuses on global spatial concepts (i.e., properties of 3D patches) while BFR addresses weaknesses of recent reconstruction methods to learn more effective local representations.
% To remedy this, , 
% By predicting multiple vectors from predefined crop positions to a consistent location across volumes, we cover all spatial concepts while learning useful features for downstream localization tasks. 
We evaluate VectorPOSE on three 3D medical image segmentation tasks, showing that it often outperforms state-of-the-art methods, especially in limited annotation settings.
% Additionally, we show improved classifier robustness and more interpretable foreground saliency.

\end{abstract}


% Main Sections
% ==========================================

\section{Introduction} \label{intro}

\begin{figure}[t!]
    \centering
    \begin{minipage}[b]{1.0\linewidth}
      \centering
      \centerline{\includegraphics[clip,trim=0.0cm 0.0cm 0.0cm 0.35cm,width=8cm]{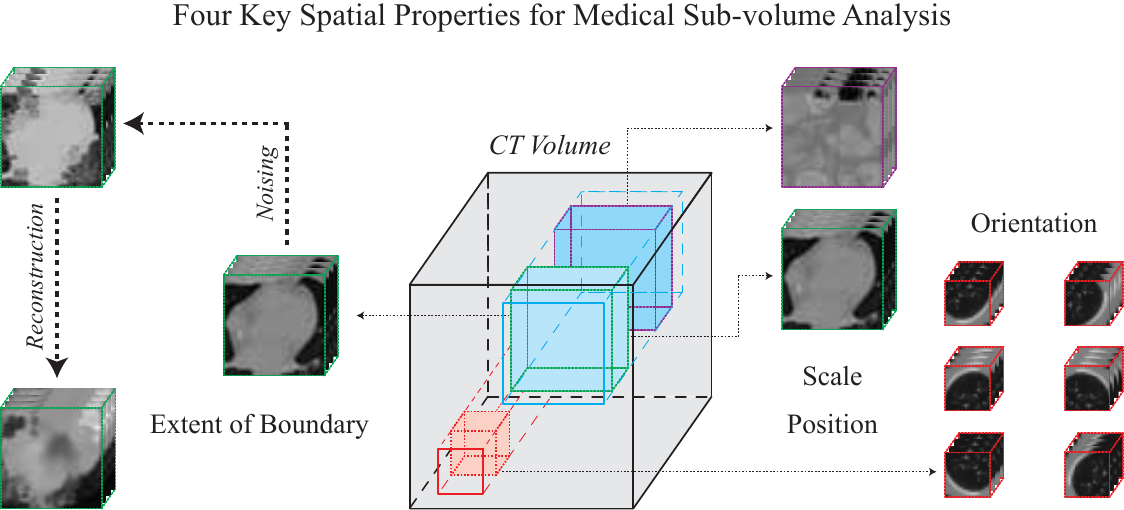}}
    %  \vspace{2.0cm}
    %  \centerline{}\medskip
    \end{minipage}
    \vspace*{-4mm}
    \caption{Spatial properties for effective 3D crop analysis. On the right, one may see the insights brought by crop position, scale, and orientation using spatial anatomical priors. On the left, one may see the importance of boundary delineation.}
    \label{fig:intro_fig}
    \vspace{-5mm}
\end{figure}

Modern 3D medical image analysis techniques have made great performance strides by extracting appearance-based features from local image patches. 
However, they fail to explicitly attend to a key 
%quality 
capability that facilitates volumetric analysis: \emph{spatial awareness}.
% with an arbitrary viewpoint 
% The importance of a 3D crop's spatial properties are best highlighted when we look from the perspective of a practitioner.
Given an arbitrary image crop (e.g., see Fig.~\ref{fig:intro_fig}), we can often leverage anatomical priors and infer a large amount of information from its position, scale, and orientation.
% The importance of spatial properties are highlighted when inspecting a random 3D patch (e.g. crops with green, purple, or red outlines in Fig. \ref{fig:res_fig}) from the perspective of a practitioner, where a surprising amount of information can be inferred with crop position, scale, and orientation (visualized on the right side of Fig. \ref{fig:res_fig}).
Crop position such as its axial window (e.g., the dashed red and blue lines in Fig.~\ref{fig:intro_fig} that span the CT volume depth) reveals which anatomical structures to expect.
% hints at valid anatomical structures that we can expect in the image.
Image zoom is commonly adjusted by practitioners, and the knowledge of a crop's scale is essential for accurate assessments of the structure size, shape, and extent. 
% normal appearance and scope of crop coverage.
Finally, crop orientation informs object poses and valid adjacent structures.
%Additionally, the spatial extent of boundaries alone allow analysis of anatomy shapes, sizes, interactions with adjacent anatomies.   
%Thus, 
From these insights, we hypothesize that \emph{networks which better predict 3D crop position, scale, and orientation can more accurately segment images with superior anatomical knowledge}.
% from strong at volumetric medical image analysis from excellent learned anatomical priors}. 

\begin{figure*}[!ht]
    \begin{minipage}[b]{1.0\linewidth}
      \centering
      \centerline{\includegraphics[width=15.5cm]{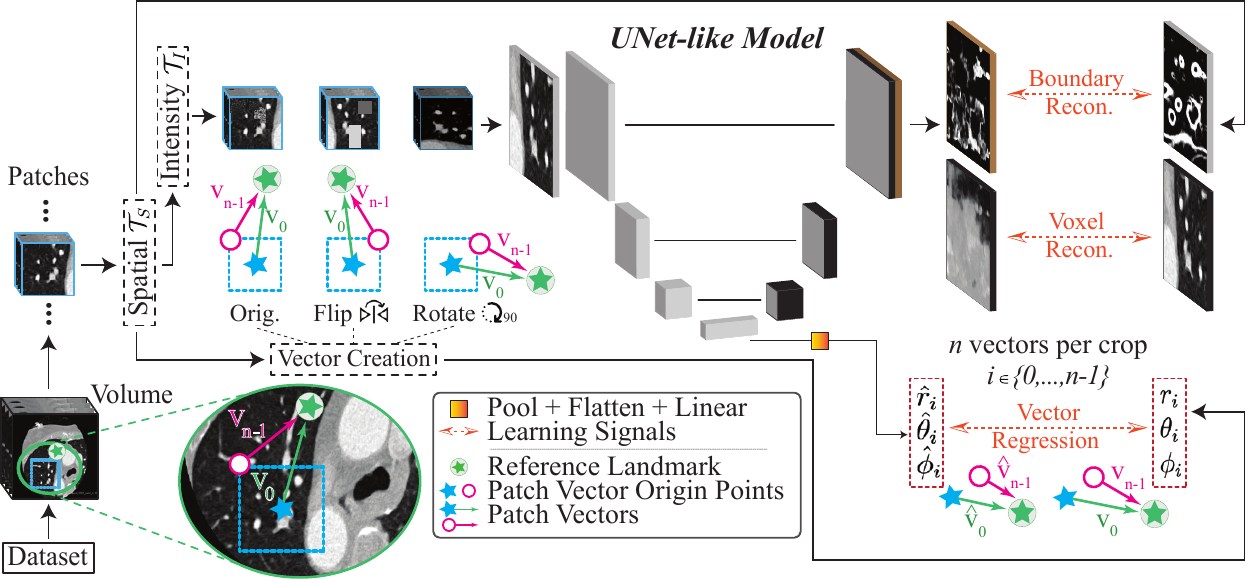}}
    %  \vspace{2.0cm}
    %  \centerline{}\medskip
    \end{minipage}
    \vspace{-7.4mm}
    \caption{\label{fig:main}An overview of our proposed VectorPOSE approach.}
    %with the Boundary-focused Reconstruction components on the top-side and Vector Prediction below.}
    \vspace{-4mm}
\end{figure*}

To distill relevant spatial information without additional annotation burden, we leverage self-supervised learning (SSL).
Recent 3D SSL methods have improved downstream task performances. They generally fall into three types.
\textbf{1) Contrastive learning} \cite{taleb20203dssl,Zheng2021HierarchicalSL,Zeng2021PositionalCL,chen2021momentumcovidclassif,chaitanya2020contrastive,Zhou2021PreservationalLI} stems from metric learning where representations of similar images (e.g., two augmented versions of an image) are encouraged to be close while features of distinct images are contrasted. 
However, many 3D medical tasks have prohibitively small datasets while requiring cubically more memory, which is incongruous with contrastive methods' reliance on large datasets, batch sizes, and models \cite{chen2020simclr}. 
% We rule out using this approach since it needs large unlabeled datasets, batch sizes, and models \cite{chen2020simple}, which is not ideal given GPU memory constraints and limited medical image dataset sizes. 
% Also, further performance degradation is expected from lower inter- and intra-class appearance variations in medical images compared to natural scene data.
\textbf{2) Reconstruction} \cite{Matzkin2020SelfsupervisedSR,Zhou2019ModelsGG,Zhang2021SARSR,Zheng2021HierarchicalSL,tao2020revisiting} has been shown to outperform supervised transfer learning on 3D data by restoring an original image from a noised or masked version.
Still, when analyzing reconstruction results in \cite{Zhou2019ModelsGG} (see the left side of Fig.~\ref{fig:intro_fig}), one can see significant losses of boundary acuity as well as tissue details. 
This may suggest an over-focus on local intensity statistics over more robust semantic or spatial understandings of the whole structures. 
%On the left of Fig. \ref{fig:res_fig}, one can see that after reconstruction (following ), regions with pixel shuffling applied "cheat" by naively predicting the average intensities of its local neighborhood.
% This may inhibit performance due to the importance that boundaries play in delineating spatial properties of structures such as shape, extent, and interaction with neighboring anatomies.
% We 
\textbf{3) Prediction}-based pretext tasks entail solving spatial ``puzzles'' such as relative position prediction \cite{taleb20203dssl}, jigsaw unscrambling \cite{taleb20203dssl,zhuang2019selfrubik1,zhu2020rubik}, or scale classification \cite{Zhang2021SARSR}. 
However, these methods are prone to short-cut learning and over-emphasize discriminative elements (e.g., patch boundaries).
Moreover, some approaches propose to learn spatial properties incompletely \cite{Zhang2021SARSR} or indirectly through disparate pretext tasks \cite{Zhou2019ModelsGG,zHaghighi2021TransferableVW}, but implicit learning is suboptimal and uncontrolled.
To our best knowledge, no approach adequately covers all key spatial properties important to 3D medical image analysis in practice.

To remedy this gap, we propose a new 3D self-supervised pretraining approach, \textbf{VectorPOSE} (POSE for \underline{P}osition, \underline{O}rientation, \underline{S}cale, and \underline{E}xtent), with two novel pretext tasks for learning improved spatial and appearance features. 
Our first pretext task, \textbf{Vector Prediction} (VP), infers vectors that originate from predefined points within a sampled input crop (e.g., crop center, corners) and terminate at the center of the volume.
% Vectors encapsulate both magnitude and direction, so predicting multiple vectors with distinct but consistent origin points effectively encodes position (with respect to the volume center), orientation (via vector angles toward a fixed reference point), and scale (where magnitude differentials between vectors represent crop extent). 
Vectors encapsulate both magnitude and direction, so predicting multiple vectors with consistent origin points effectively encodes position (with respect to the terminal point), orientation (via vector angles to the terminal point), and scale (where differentials between predicted vector origins represent crop extent). 
To learn effective local features and address boundary degredations in reconstruction, our second pretext task, \textbf{Boundary-Focused Reconstruction} (BFR), conducts: 1) \textit{boundary reconstruction} where extracted edges of the patch are explicitly predicted, and 2) \textit{voxel reconstruction} where original intensity values are predicted with a loss that better preserves boundaries.
% learning learns boundary extent information of anatomical structures through 2 prediction tasks: 1) voxel reconstruction where original intensity values are predicted, and 2) boundary reconstruction where extracted edges of the original crop are predicted.
%
% pretext task where the model predicts vectors that originate from predefined key-points within a given input patch  and terminate at the center of the original volume. 
% By predicting multiple vectors, we effectively cover all the three key spatial properties: position is predicted with respect to the volume center, orientation is computed from vector angles, and scale is understood since the extent of the crop needs to be known in order to accurately point vectors from 2 corners of the crop to the terminal point.
%
% Next, we propose a \textbf{Boundary-Focused Reconstruction} (BFR) task that learns boundary extent information of anatomical structures through 2 prediction tasks: 1) voxel reconstruction where original intensity values are predicted, and 2) boundary reconstruction where extracted edges of the original crop are predicted.
To demonstrate efficacy, we pretrain and evaluate on three 3D semantic medical image segmentation tasks (i.e., CT cardiac, CT abdominal, MR prostate).
VectorPOSE generally outperforms state-of-the-art single- and multi-task methods while exhibiting superior data efficiency compared to known contrastive learning methods.

% Comparing to state-of-the-art single- and multi-task self-supervised approaches, our framework generally outperforms them while avoiding burdens from multi-task and adversarial trainings.

% and evaluate its performance by fine-tuning on a whole heart CT segmentation task and an abdominal organ CT segmentation task (both included in the four pretraining sets).

%\\~\\
%\noindent
Our \textbf{contributions} are summarized below.
\begin{enumerate}
    \item We propose a self-supervised approach, VectorPOSE, that places a holistic emphasis on three key spatial properties (position, orientation, scale) while synergizing learned global information with local, boundary-aware cues.
    % effective pretext task, Vector Prediction (VP), that succinctly encapsulates three key spatial properties for effective volumetric analysis along with a boundary-focused reconstruction (BFR) task for appearance feature learning.
    \item As a cornerstone to spatial understanding, we introduce a novel Vector Prediction (VP) pretext task that succinctly encapsulates key spatial properties and is both effective and extensible. Further, we propose the Boundary-Focused Reconstruction (BFR) pretext task to address pitfalls of recent works and learn improved local features.
    % self-supervised framework  that learns both task-specific and general representations of 3D medical images, is simple to train, and easily fits within general deep learning pipelines. 
    \item VectorPOSE is comprehensively evaluated on two 3D CT segmentation datasets (MMWHS \cite{zhuang2016multimmwhs,zhuang2013challengesmmwhs} \& BCV \cite{bcv2015}) and one 3D MR prostate dataset \cite{Antonelli2022TheMSD}. Not only does it outperform recent self-supervised methods in the majority of settings, it also exhibits improved data efficiency without additional training time.
    % \item We (will) release our implementations and pretrained model parameters to facilitate research in the area of self-supervised 3D medical image analysis.
\end{enumerate}

\vspace{-2mm}
\section{Methodology} \label{method}

Before detailing the proposed pretext tasks, we first overview our general pipeline and its major components.
% major components and general data-flows.
Following established self-supervised methods, we pretrain a randomly initialized neural network consisting of an encoder $E$, a decoder $D$, the prediction head $D_{\textit{vp}}$ for Vector Prediction (VP), and the prediction head $D_{\textit{bfr}}$ for Boundary-Focused Reconstruction (BFR).
% Predictions for the two main pretext tasks, Vector Prediction (VP) and Boundary-focused Reconstruction (BFR), are outputted by prediction heads $D_{v}$ and $D_{b}$, respectively.
Note that $D_{\textit{vp}}$ employs global information and processes pooled \& flattened output features from $E$, while $D_{\textit{bfr}}$ utilizes $D$'s output that has the same resolution as $E$'s input.
To pretrain $\{E, D, D_{\textit{vp}}, D_{\textit{bfr}}\}$ with VectorPOSE (see Fig.~\ref{fig:main}), we input patches $x^{i(j)}, x^{i(k)}, \ldots$ from the $i$th image $X^i$ of an unlabeled dataset $\mathcal{D}_u$.
Stochastic spatial augmentations $\mathcal{T}_S$ (e.g., flipping or rotating along each axis) and intensity noising $\mathcal{T}_I$ (e.g., local pixel shuffle, masking, intensity shifting) are applied to all input patches independently and in order. 
The targets for VP \& BFR are computed after $\mathcal{T}_S$ is applied.
% $(X^i, Y^i) \in \mathcal{D}$
% We denote the images and masks from a dataset as .
% For network components, $E$ is the encoder, $D$ is the main decoder, $D_{v}$ is the predictor head for VP, and $D_{b}$ indicates the predictor for BFR.  
After pretraining, 
%heads 
$D_{\textit{vp}}, D_{\textit{bfr}}$ are discarded while $E, D$ are transferred and fine-tuned on a downstream target task. 

\vspace{-2mm}
\subsection{Vector Prediction (VP)}

The objective of VP is to distill position, scale, and orientation information from crops by predicting multiple vectors which all point to one standardized location in all dataset volumes (e.g., volume center or center of a chosen object).
We refer to this point as the \textit{reference landmark}, $(x^{i(j)}_{\dashv},y^{i(j)}_{\dashv},z^{i(j)}_{\dashv})$, which denotes the landmark point for crop $x^{i(j)}$ from image $X^i$.
% These points are determined for every volume and serve as the terminal points for all vectors within an image. 
% Spatial focus about the reference point allows models to learn useful priors by exploiting anatomical consistencies.
% Spatial focus with respect to exploit the anatomical consistency of structures across images to learn useful priors. 
Reference landmarks across images have anatomical correspondences from which we learn useful priors and structural consistencies.
In many 3D tasks, reference landmarks can simply be taken as the center of volumes given rough image alignments.
In others, positions of landmarks can be extracted in an unsupervised manner (e.g., center of the left lung as a landmark after lung masks are automatically extracted \cite{zhao2003automaticlung}).
To regularize the reference landmark, we randomly jitter coordinates by $\eta$\% of each image dimension's total length.

After image landmarks are obtained, a vector's origin point concretely defines distance (via its magnitude) and position (via angles) to the reference.
For a sampled crop $x^{i(j)}$, we define $n$ vector origin points as $(x^{i(j)}_{0\vdash},y^{i(j)}_{0\vdash},z^{i(j)}_{0\vdash})$, $\ldots$, $(x^{i(j)}_{n-1\vdash},y^{i(j)}_{n-1\vdash},z^{i(j)}_{n-1\vdash})$.
In practice, we set these origin points in consistent positions for all crops (e.g., $m$=$0$ at patch center, $m$=$1$ at patch top-right corner, etc.).
% , canonical
% We assume that this location 
% from patches through a succinct multi-vector prediction loss.
% A vector's start point lies on the same canonical position within each patch (e.g., top left corner of a patch extracted from the original volume) while the end point is fixed to be the axial-adjusted volume center.
% Concretely, if a crop $x^{i(j)}_{dn}$ has center coordinates of $(x^{i(j)}, y^{i(j)}, z^{i(j)})$ and its original volume has dimensions of $(D, H, W)$, then the end point for all vectors from the $j$-th crop in volume $i$, $x^{i(j)}_{dn}$, is $(x^{i(j)}, H/2, W/2)$.
% We choose the axial-adjusted end point because volumes within a dataset often have largely varying depths so we want the prediction target to be mostly consistent. 
% Also, we only train this task on downstream data because different tasks often have different axial coverage (e.g., MMWHS centers the volumes around the heart and BCV centers in the middle of torsos), so predicting widely varying reference points will be harmful to feature learning.
Mathematically, predicting a single vector is sufficient to describe position and orientation, while two are enough to also cover scale.
However, we empirically find that $n$=$9$ vectors (vectors originating from the patch center and 8 corners) improve feature learning (see Section \autoref{exp}, Tab. \ref{tab:ablation}).
% This probably allows the model to learn from a larger variety of visual localities and manages cases where large portions of patches may contain uninformative contents.
% edge cases where patches lie on uninformative local regions so other corner vectors can compensate. 

Thus, all vectors belonging to crop $x^{i(j)}$ are defined as $v_m^{i(j)}$ ($m =0, \ldots, n-1$) with origin point $(x^{i(j)}_{m\vdash},y^{i(j)}_{m\vdash},z^{i(j)}_{m\vdash})$ and terminal point $(x^{i(j)}_{\dashv},y^{i(j)}_{\dashv},z^{i(j)}_{\dashv})$. 
The terminal points are identical for all crops from the same image.
Below we omit ``$i(j)$'' for readability.
For prediction, we reparameterize vectors into spherical coordinates: 
let $(x_m,y_m,z_m) = (x_{\dashv} - x_{m\vdash}, y_{\dashv} - y_{m\vdash}, z_{\dashv} - z_{m\vdash})$; so, $v_m = $ ($r_m$, $\theta_m$, $\rho_m$), where $r_m =$ $ \sqrt{x_m^2 + y_m^2 + z_m^2}$, $\theta_m = \arccos(z_m / r_m)$, and $\phi_m = \arctan(y_m / x_m)$ with constraints $r \in [0, \inf)$, $\theta \in [0, \pi]$, and $\phi \in [-\pi, \pi]$.
The model regresses the $n$ normalized spherical coordinates of a crop via the VP loss:
% \vspace*{-3mm}
% \begin{equation} \label{eq:1}
%     \begin{split}
%     \mathcal{L}_{\textit{vp}} &= \frac{1}{n} \sum_{m=0}^{n-1} \left(||\frac{r_m}{R} - \sigma(\hat{r}_m)||_1
%     + ||\frac{\theta_m}{\pi} - \sigma(\hat{\theta}_m)||_1 + \right.\\ 
%     &\min(||\frac{\phi_m}{\pi} - \tanh(\hat{\phi}_m)||_1,
%          ||\frac{\phi_m + 2\pi}{\pi} - \tanh(\hat{\phi}_m)||_1,\\
%           &||\left.\frac{\phi_m - 2\pi}{\pi} - \tanh(\hat{\phi}_m)||_1)  \right),
%     \end{split}
% \end{equation}
% \vspace*{-3mm}

% NICK'S EDITTTTTTTTTTTTTTTTT
\small{
\vspace*{-3mm}
\begin{equation} \label{eq:1}
\begin{aligned}[b]
    \mathcal{L}_{\textit{vp}} = &\frac{1}{n} \sum_{m=0}^{n-1} \left(||\frac{r_m}{R} - \sigma(\hat{r}_m)||_1
    + ||\frac{\theta_m}{\pi} - \sigma(\hat{\theta}_m)||_1 + \right.\\ 
    \min(&||\frac{\phi_m}{\pi} - \tanh(\hat{\phi}_m)||_1,
         ||\frac{\phi_m + 2\pi}{\pi} - \tanh(\hat{\phi}_m)||_1,\\
          &||\left.\frac{\phi_m - 2\pi}{\pi} - \tanh(\hat{\phi}_m)||_1)  \right)
\end{aligned}
\end{equation}
\vspace*{-3mm}
}

% Concretely, given the a vector from patch $x^{i(j)}$
% We define the $k$-th 3D vector of a patch using its spherical coordinate representation, $v_k = $ ($r_k$, $\theta_k$, $\rho_k$).
% We can obtain these given a start point $(x_s, y_s, z_s)$ and end point $(x_e, y_e, z_e)$ by first converting them into a positional vector represented as the coordinates $(x_e - x_s, x_e - y_s, x_e - z_s)$. 
% Then, $r = \sqrt{x^2 + y^2 + z^2}$, $\theta = arccos(z / r)$, and $\phi = arctan(y / x)$ with $r \in [0, \inf)$, $\theta \in [0, \pi]$, and $\phi \in [-\pi, \pi]$.
% For $K$ total vectors to predict from a crop $x^{i(j)}_{dn}$, the output prediction logits, $E(D_v(x^{i(j)}_{dn}))$, is of size $3*K$.

% Given the logits vector $\hat{y} = E(D_v(x^{i(j)}_{dn}))$ for crop $x^{i(j)}_{dn}$, we define the VP loss as:
    % \\[-2em]
    % %\begin{equation}
    % \begin{multline}
    %         \resizebox{.93\hsize}{!}{$\mathcal{L}_{vp}^{i(j)} = \frac{1}{K} \sum_{k=0}^{K-1} ||\frac{r_k}{R} - \sigma(\hat{y}[3k])||_1 
    %         + ||\frac{\theta_k}{\pi} - \sigma(\hat{y}[1 + 3k])||_1
    %         + ||\frac{\phi_k}{\pi} - \Gamma(\hat{y}[2+3k])||_1$},
    % \end{multline}
    %\end{equation}
\noindent
where $\sigma$ is the sigmoid function, $R$ is the radius of a sphere circumscribing the image volume, and $\hat{r}_m$, $\hat{\theta}_m$, $\hat{\phi}_m$ are logits from $E \circ D \circ D_{\textit{vp}}$ (see the bottom right of Fig.~\ref{fig:main}).
To address the $\phi$ angles that are physically close but distant in the parameter space (e.g., $-179\degree$ \& +179$\degree$), we take the minimum among targets $\phi_m, \phi_m - 2\pi, \phi_m + 2\pi$ (see 
%bottom of 
Eq.~(\ref{eq:1})).
Also, note that when crops undergo spatial augmentations (e.g., flipping, rotations), 
the index of each vector's post-transform origin corresponds with the origin before transforming.
% % $\theta_m$ \& $\phi_m$ follow the directional transformation of the crop.
% Also, for sake of readability, we omit the details of handling angles that are spatially close but distant in parameter space (e.g. -179$\degree$ \& +179$\degree$). 
% We refer readers to our code for further details.

By encapsulating the three key spatial properties into an interdependent task, we prevent training instability and feature interference from independent predictions like when classifications were used \cite{Zhang2021SARSR}. 
Additionally, vectors represent spatial properties in a continuous space which is a generalized extension of discrete rotation or scale prediction \cite{zhu2020rubik}.
% \begin{itemize}
%     \item General Story: (1) introduce pretext task (single vector) that covers position and orientation.(2) show this doesn't cover scale because of the concentric patch problem. then propose to use vectors from all sides. (3) drill in all the benefits of prediction all 9 vectors (1. more heterogeneous learning signal, 2. learning dilution in an edge case where patch is close to the center \& edges are on all 4 sides of volume center, (4) justify the loss that we chose to use. 
% \end{itemize}

\vspace{-2mm}
\subsection{Boundary-Focused Reconstruction (BFR)}
Complementing the global features from VP, two appearance-focused losses are employed in BFR. 
To address previous issues of boundary degradation and reduced texture acuity, we first switch the reconstruction criterion from $L_2$ loss (as used in \cite{Zhou2019ModelsGG}) to $L_1$ loss, which has been shown to improve both reconstuction quality and boundary accuracy \cite{tao2020revisiting}. 
%to alleviate blurring during voxel reconstruction like in \cite{Zhou2019ModelsGG}.
To further emphasize anatomical delineation, we add an additional boundary reconstruction task where boundaries are extracted via a 3D Scharr edge detector.
We select the Scharr transform since it exhibits superior rotational invariance compared to other filters like Sobel.
In conjunction with strong augmentations such as texture noising (e.g., local pixel shuffle) and masking (e.g., in-painting or out-painting), explicit edge reconstruction facilitates the learning of general anatomical shapes and boundary extents.
%  when contextual clues are heavily occluded.
% By explicitly tasking the model to predict boundaries, we facilitate the learning of both noised object boundaries and boundaries that require contextual clues for proper extrapolation after masking.
%
% To better distill spatial extent awareness during pretraining, we add a boundary reconstruction task where a 3D Scharr edge detector extracts anatomical boundaries and the model is tasked with reconstructing the original edges from a noised image.
The overall BFR loss on a crop is defined as:
\vspace{-2.2mm}
\begin{equation}
    \mathcal{L}_{\textit{bfr}} = \frac{1}{n} \sum_{m=0}^{n-1} 
    (|| {y}_m^v - \sigma(\hat{y}^v_m) ||_1 + 
    \alpha || y_m^b - \sigma(\hat{y}^b_m) ||_1),
\vspace{-2.2mm}
\end{equation}
where ${y}_m^v$ and $y_m^b$ are the targets of the voxel and boundary reconstruction tasks,
%respectively, 
% $\hat{r}_v^{i(j)} = \sigma(E(D_r(x^{i(j)}))[0])$ \& $\hat{r}_v^{i(j)} = \sigma(E(D_r(x^{i(j)}))[1])$ are the voxel \& boundary reconstruction predictions, respectively.
$\hat{y}^v_m$ and $\hat{y}^b_m$ (from $E \circ D \circ D_{\textit{bfr}}$) are the voxel and boundary logits, respectively. 
$\alpha$ is a scaling term to address the small proportion of boundaries in crops.

\vspace{-2mm}
\subsection{Overall Loss and Implementation}

The overall VectorPOSE loss for an image crop is:
\vspace{-2mm}
\begin{equation}
    \mathcal{L} = \lambda\mathcal{L}_\textit{bfr} + (1-\lambda)\mathcal{L}_\textit{vp}.
\vspace{-2mm}
\end{equation}

\noindent 
For our UNet-like model, we employ a 3D version of ResNet-50 \cite{He2016DeepRLResNet} as the encoder and attach it to a light decoder with bilinear upsampling layers \& additions for feature fusion.
For $D_{\textit{vp}}$, the lowest encoder output 
%resolution 
is pooled, flattened, and processed through a 2-layer MLP with 256 hidden dimensions, while $D_{\textit{bfr}}$ uses a single 1x1 convolution. 
We use $m$=$9$ vectors per patch with a jitter $\eta$=$5$\%, and train with $\alpha$=$5$ \& $\lambda$=$0.5$.

\newcommand{\ra}[1]{\renewcommand{\arraystretch}{#1}}
\setlength{\tabcolsep}{4pt}{
\begin{table*}[!h]
    \centering
    \ra{1.1} 
    \begin{tabular*}{0.92\linewidth}{@{}lllllllllllllll@{}}
        \toprule
        \multirow{2}{*}[-0.15em] {\hspace{1.5pt} \textbf{Method}} &
        \multicolumn{4}{c}{CT MMWHS} & {} & \multicolumn{4}{c}{CT BCV} & {} & \multicolumn{4}{c}{MRI MSD-Prostate}\\
        \addlinespace[0pt]
        \cmidrule(){2-5} \cmidrule(){7-10} \cmidrule(){12-15}
        % {} & Accuracy & Recall & Precision & F_1  && 
        {} & 10\% & 25\% & 50\% & 100\% &&
        10\% & 25\% & 50\% & 100\% && 10\% & 25\% & 50\% & 100\% \\
        \midrule
        Random Init.  
            & 79.31 & 87.29 & 89.15 & 90.70 && 
            30.06 & 58.48 & 66.23 & 75.81 &&
            40.05 & 58.78 & 69.25 & 74.24 \\
        Models Genesis \cite{Zhou2019ModelsGG}  
            & 80.05 & 88.14 & 90.54 & 91.05 && 
            32.03 & 58.80 & 66.87 & 75.79 &&
            \underline{44.00} & 60.89 & 69.55 & \underline{75.86} \\
        SAR \cite{Zhang2021SARSR}             
            & \underline{81.48} & \underline{88.57} & 90.81 & 91.14 && 
            \underline{35.63} & \underline{60.29} & 67.52 & \textbf{76.77} &&
            42.53 & 61.02 & 70.16 & 74.62 \\
        Rubik++ \cite{tao2020revisiting}         
            & 81.08 & 88.40 & \underline{90.94} & \underline{91.23} && 
            33.28 & 59.83 & \underline{67.72} & \underline{76.65} &&
            43.70 & \underline{61.97} & \textbf{71.35} & 75.81 \\
        PGL \cite{Xie2020PGLPL}             
            & 79.76 & 87.50 & 89.48 & 90.57 && 
            30.99 & 58.56 & 66.39 & 75.98 &&
            41.36 & 59.08 & 69.97 & 74.08 \\
        MoCo \cite{chen2021momentumcovidclassif}             
            & 80.22 & 87.94 & 89.60 & 91.05 && 
            31.89 & 58.95 & 66.91 & 76.01 &&
            42.07 & 60.69 & 70.17 & 74.31 \\
            \hline
        %Hierarchical\footnote{Trained on 2D slices.} \cite{Zheng2021HierarchicalSL} 
        %    & 99.99 & 99.99 & 99.99 & 99.99 && 99.99 & 99.99 & 99.99 & 99.99 \\
        VectorPOSE (Ours)  
            & \textbf{83.30} & \textbf{89.43} & \textbf{91.27} & \textbf{91.60} && 
            \textbf{37.88} & \textbf{61.61} & \textbf{68.01} & 76.53 &&
            \textbf{46.35} & \textbf{63.14} & \underline{71.22} & \textbf{75.97} \\
        \multicolumn{1}{r}{\textit{p-value}} 
            & \textit{.0335} & \textit{.0286} & \textit{.0437} & \textit{.0564} && 
            \textit{.0179} & \textit{.0156} & \textit{.0388} & \textit{.0593} &&
            \textit{.0082} & \textit{.0098} & \textit{.0571} & \textit{.0681} \\    
            
        \bottomrule
    \end{tabular*}
    \vspace*{-2mm}
    \caption{\label{tab:res_main}Class-averaged Dice score (\%) comparisons vs.~recent self-supervised 3D medical image segmentation methods. Entries that are bolded \& underlined are the \textbf{best} \& \underline{second-best} scores, respectively. p-values are obtained from independent t-tests of the best and second-best scores.}
    \vspace*{-1mm}
\end{table*}
}

\setlength{\tabcolsep}{3.4pt}{
\begin{table}[!t]
    \centering
    \vspace*{0mm}
    \ra{1.2} 
    \begin{tabular*}{1.0\linewidth}{@{}r|cccccc@{}}
        \toprule
        Voxel Rec.          & $\mathcal{L}_1$
                            & $\mathcal{L}_1$ 
                            & $\mathcal{L}_1$ 
                            & $\mathcal{L}_1$ 
                            & $\mathcal{L}_1$
                            & $\mathcal{L}_1$\\
        Bound. Rec.       &  
                            & $\mathcal{L}_1$ 
                            & $\mathcal{L}_1$ 
                            & $\mathcal{L}_1$ 
                            & $\mathcal{L}_1$
                            & $\mathcal{L}_1$\\
        \midrule
        Center Vector       &  &
                            & \cmark
                            & \cmark
                            & \cmark
                            & \cmark \\
        Corner Vector(s) &  & 
                            &
                            & 1 
                            & 4 
                            & 8 \\
        \bottomrule 
        \vspace*{3pt}
        Dice (\%)           & 80.16 
                            & 80.80
                            & 81.37
                            & 82.29
                            & 82.75
                            & \textbf{83.30} \\
    \end{tabular*}
    \vspace*{-4mm}
    \caption{\label{tab:ablation}Ablations on the proposed components on 10\% MMWHS. 
    The top two rows contain appearance-based components while the middle two rows have VP settings.}
    \vspace{-3mm}
    % (e.g., loss for voxel reconstruction, loss for boundary reconstruction, and edge extraction algorithm), the middle two rows indicate various Vector Prediction settings (e.g., the presence of the center vector, and the number of corner vectors to predict), and the bottom row shows the percentage Dice scores with the right-most entry being equivalent to our final method.}
\end{table}
}

\vspace{-2mm}
\section{Experiments \& Results} \label{exp}

We implemented all experiments in PyTorch and trained on NVIDIA Tesla P100s (16GB VRAM).
For pretraining, volume intensities were normalized between 0 and 1, uninformative regions were excluded, (96, 96, 96) crops were sampled, and augmentations in \cite{Zhou2019ModelsGG} were applied.
We trained using AdamW with a $0.0002$ learning rate \& $0.0001$ weight decay for 300 epochs ($\sim$21 hours) with a batch size of 12. 
For fine-tuning, we normalized volume intensities between 0 and 1, sampled (64, 128, 128) crops, and applied random flipping, brightness, gamma, and blurring.
We optimized using AdamW with a learning rate $0.001$ and a weight decay $0.001$ for 200 epochs ($\sim$14 hours) with a batch size of 4 over 8 runs. 
For evaluation, we used the average Dice-Sørensen Coefficient across foreground classes.

\vspace{-2mm}
\subsection{Datasets}

All the datasets are split using a 6:2:2 training:validation:test ratio.
Models are pretrained using the full training sets and fine-tuned using 10\%, 25\%, 50\%, or 100\% of the training data (if the number of training samples is not divisible, the floor function is applied).

\textbf{CT MMWHS}
% (\textbf{M}ulti-\textbf{M}odality \textbf{W}hole \textbf{H}eart \textbf{S}egmentation)
\cite{zhuang2013challengesmmwhs,zhuang2016multimmwhs} provides 20 annotated CT volumes segmenting seven cardiac structures (left ventricle, left atrium, right ventricle, right atrium, myocardium, ascending aorta, and pulmonary artery). 
\textbf{CT BCV} 
% \textbf{B}eyond the \textbf{C}ranial \textbf{V}ault 
\cite{bcv2015} is an abdominal organ segmentation challenge with 30 annotated CT volumes and 13 classes.
\textbf{MRI MSD-Prostate} 
% \textbf{M}edical \textbf{S}egmentation \textbf{D}ecathlon 
\cite{Antonelli2022TheMSD} contains 32 labeled MRI (T2, ADC) volumes with two foreground classes.

\subsection{Results and Discussion}

% We elect to compare our method with the most recent multi-task self-supervised methods since they empirically outperform single task methods \cite{Taleb20203DSM} or already contain the single task within their frameworks. 
% For those that did not release code, we faithfully implemented their methods to the best of our ability, and use the recommended hyperparameters mentioned in their original works.
For fair comparisons, all the methods were pretrained and fine-tuned on the same data splits with the same number of iterations, and allowed an equal budget for hyperparameter tuning. 
The performances on the three downstream datasets are summarized in Table \ref{tab:res_main}. 
% and qualitative results in Fig. \ref{fig:supp1}.

From our experimental results, we make the following observations.
\textbf{1)} Our approach generally \textit{outperforms recent state-of-the-art multi-task methods} and handily beats training from scratch, which supports our hypothesis that more spatially aware models yield more effective representations of 3D crops. 
More details on the effects of spatially-aware components are discussed in Section \autoref{Ablations}.
% We further discuss the improvements that spatially-aware pretraining invokes in Section \ref{Ablations}. 
\textbf{2)} The \textit{fewer the annotations, the bigger the performance improvements}. This is a desirable property for tasks with many biomedical objects with extremely limited labels.
Our outperformance over other methods that also use reconstruction \cite{Zhou2019ModelsGG,Zhang2021SARSR,tao2020revisiting} shows both the efficacy of our spatial pretext task and superior data efficiency.
% , our spatial pretext task proves to effectively complement reconstruction, enhance representations, and exhibit better data efficiency.
% our framework is able to learn more effective features from spatially-focused pretext tasks. 
% \textbf{3)} Our method handily \textit{beats training from scratch} and shows its potential as a reasonable replacement to random initialization methods given that both require no annotations.
\textbf{3)} Both our \textit{encoder and decoder are initialized with multi-scale and semantically diverse tasks}. 
Many methods only learn encoder features while suboptimally neglecting decoder parameters (e.g., \cite{Xie2020PGLPL}).
Additionally, the decoder receives heterogeneous learning signals from both local-focused BFR and global, spatially-aware VP.
This allows for each segment of the model to learn relevant and transferable features and serves as a natural regularizer that prevents overfitting to any single task.
In fact, we notice that joint training leads to synergistic improvements.
\textbf{4)} Lesser performance from PGL \& MoCo may indicate \textit{innate disadvantages in using metric learning techniques with 3D radiology tasks}. 
We hypothesize that this is in part due to the fact that PGL only uses positive samples which inhibits feature learning.
The small amount of pretraining images also limits effectiveness.
Given that many 3D medical images are highly structured and prior-dependent, we can more effectively learn these priors through spatial prediction tasks that holistically cover multiple key properties. \\[-1.5em]

\subsection{Ablations} \label{Ablations}

Here, we explore the contributions of each individual component in our method. 
Notably, we intend to test our hypothesis about the importance of explicitly learning spatial parameters and compare its efficacy against other proposed methods (e.g., reconstruction).
In Table \ref{tab:ablation}, one can see $\sim$0.6\% improvement after adding the boundary awareness task, which supports the benefit of explicitly predicting boundaries over voxel reconstruction alone.
% existing reconstruction methods prioritize naive local intensities over structural extents and that predicting these boundaries is helpful for feature learning.
Next, we see that with increasing $m$ (the number of vectors per crop), performance notably rises. 
This supports our hypothesis of using spatial awareness as an effective feature-learning principle. 
% proxy task for learning helpful spatial features and is able to transfer those priors to downstream segmentation.

% Other Sections: Conclusion, Ethical, Refs
% ==========================================

\section{Conclusions}

We proposed VectorPOSE, a new self-supervised approach for 3D medical image segmentation with two novel pretext tasks: 1) Vector Prediction (VP) to address the lack of spatial understanding (i.e., position, scale, orientation) in existing works; 2) Boundary-Focused Reconstruction (BFR) for improving the understanding of anatomical structure extents through boundary regression.
After evaluation on three 3D semantic segmentation datasets, VectorPOSE's general outperformance over state-of-the-art methods shows that spatial understanding is pertinent for feature learning and may be a promising avenue for future research progress.  

% that specialize in learning effective spatial as well as appearance features.
% Our first pretext task, Vector Prediction (VP), was motivated by our observation that existing methods neglect three key spatial properties that are important to 3D crop analysis: position, scale, and orientation.
% Our Boundary-focused Reconstruction (BFR) task aims to address shortfalls in noised reconstruction pretraining by focusing more on boundary extents and, consequently, more accurate spatial bounds of anatomical structures in patches. 
% We evaluated our method on three 3D semantic segmentation datasets (two CT \& one MR) and demonstrated general outperformance of current state-of-the-art multi-task pretraining methods.
% We also conduct an ablation study on the contributions of each of our proposed components and show the positive impact that spatial prediction has on learning effective features. 

\section{Compliance with ethical standards}
\label{sec:ethics}
\vspace{-2mm}
This research was conducted retrospectively using open access human subject data by three publicly available datasets \cite{zhuang2013challengesmmwhs,bcv2015,Antonelli2022TheMSD}. 
Ethical approval was not required as confirmed by the licenses attached with the open access datasets.
\vspace{-3mm}

{
    \bibliographystyle{IEEEbib_abbrev}
    \bibliography{refs}
}

%% ------------------------------------------------------------ %%
%% ------------------------------------------------------------ %%
%% ------------------------------------------------------------ %%
%% ------------------------------------------------------------ %%
%% ------------------------------------------------------------ %%

\end{document}